\documentclass[conference]{IEEEtran}
\usepackage{graphicx}
\usepackage{shortcuts}
\usepackage{xspace}
\usepackage{code}
\usepackage{cite}
\usepackage{subcaption}
\usepackage{footnote}
\usepackage{mathtools}
\usepackage{url}
\usepackage{chngcntr}
\usepackage{listings, color, caption}
\usepackage{threeparttable}
\usepackage{setspace}
\usepackage{multirow}
\usepackage{multicol}
\usepackage{pifont}
\usepackage{soul}
\usepackage{boxedminipage}

\usepackage{algorithm,algorithmicx}
\usepackage[noend]{algpseudocode}

\usepackage{comment}
\graphicspath{ {images/} }
\usepackage{amsfonts}
\usepackage{mathrsfs}
\usepackage{amssymb}
\usepackage{amsthm}

\usepackage{multirow}
\usepackage{enumitem}
\usepackage{tabularx}
\usepackage{booktabs}

\usepackage[breaklinks=true]{hyperref}

\definecolor{gray}{rgb}{0.4,0.4,0.4}
\definecolor{darkblue}{rgb}{0.0,0.0,0.6}
\definecolor{cyan}{rgb}{0.0,0.6,0.6}

\lstdefinelanguage{XML}
{
  morestring=[b]",
  morestring=[s]{>}{<},
  morecomment=[s]{<?}{?>},
  stringstyle=\color{black},
  identifierstyle=\color{darkblue},
  keywordstyle=\color{cyan},
  morekeywords={xmlns,version,type}
}
\lstdefinelanguage{JavaScript}{
    keywords={typeof, new, true, false, catch, function, return, null, catch, switch, var, if, in, while, do, else, case, break},
    keywordstyle=\color{blue}\bfseries,
    ndkeywords={class, export, boolean, throw, implements, import, this},
    ndkeywordstyle=\color{green}\bfseries,
    identifierstyle=\color{black},
    sensitive=false,
    comment=[l]{//},
    morecomment=[s]{/*}{*/},
    commentstyle=\color{purple}\ttfamily,
    stringstyle=\color{blue}\ttfamily,
    morestring=[s]{`}{'},
}
\definecolor{lightgray}{rgb}{.9,.9,.9}
\definecolor{darkgray}{rgb}{.4,.4,.4}
\definecolor{purple}{rgb}{0.65, 0.12, 0.82}
\newcommand{\RN}[1]{%
  \textup{\uppercase\expandafter{\romannumeral#1}}%
}

\algrenewcommand\algorithmicindent{0.7em}%

\title{A Depression Detection Method Based on Multi-Modal Feature Fusion Using Cross-Attention}

\author{
    \IEEEauthorblockN{Shengjie Li\IEEEauthorrefmark{2}, Yinhao Xiao\IEEEauthorrefmark{1}
        }

    \IEEEauthorblockA{\IEEEauthorrefmark{2}School of Information Science, Guangdong University of Finance and Economics, Guangzhou, China.
    \\ lsj@student.gdufe.edu.cn}

    \IEEEauthorblockA{\IEEEauthorrefmark{1}School of Information Science, Guangdong University of Finance and Economics, Guangzhou, China.\\
                      \textit{20191081@gdufe.edu.cn}\\
                      \textsuperscript{*}Corresponding Author}
}

\begin{document}
\maketitle
\begin{abstract}
Depression, a prevalent and serious mental health issue, affects approximately 3.8\% of the global population. Despite the existence of effective treatments, over 75\% of individuals in low- and middle-income countries remain untreated, partly due to the challenge in accurately diagnosing depression in its early stages. This paper introduces a novel method for detecting depression based on multi-modal feature fusion utilizing cross-attention. By employing MacBERT as a pre-training model to extract lexical features from text and incorporating an additional Transformer module to refine task-specific contextual understanding, the model's adaptability to the targeted task is enhanced. Diverging from previous practices of simply concatenating multimodal features, this approach leverages cross-attention for feature integration, significantly improving the accuracy in depression detection and enabling a more comprehensive and precise analysis of user emotions and behaviors. Furthermore, a Multi-Modal Feature Fusion Network based on Cross-Attention (MFFNC) is constructed, demonstrating exceptional performance in the task of depression identification. The experimental results indicate that our method achieves an accuracy of 0.9495 on the test dataset, marking a substantial improvement over existing approaches. Moreover, it outlines a promising methodology for other social media platforms and tasks involving multi-modal processing. Timely identification and intervention for individuals with depression are crucial for saving lives, highlighting the immense potential of technology in facilitating early intervention for mental health issues.
\end{abstract}

\section{Introduction}
\label{sec:introduction}
Depression(major depressive disorder) is a common and serious mental disorder that negatively affects how you feel, think, act, and perceive the world\cite{whatisdepression}.According to estimates from the World Health Organization (WHO), around 3.8\% of the global population is affected by depression, including 5\% of adults—comprising 4\% of men and 6\% of women—and 5.7\% of adults aged 60 years and older. Approximately 280 million people worldwide suffer from depression, with women having a roughly 50\% higher chance of developing the disorder compared to men. More than 10\% of pregnant women and those who have recently given birth globally also experience depression. Each year, over 700,000 individuals die by suicide, making it the fourth leading cause of death among individuals aged 15 to 29. Despite the availability of effective treatments for mental illnesses, over 75\% of individuals in low- and middle-income countries do not receive any form of treatment\cite{depressivedisorder}. The challenge in making accurate diagnoses during the early stages of depression contributes to a substantial number of patients being unable to access timely diagnosis and care.

With the widespread adoption of the internet and social media, these platforms have gradually evolved into a new window for studying mental health, particularly in identifying early signs of depression\cite{fusar2023lived}. The digital footprints users leave online, such as posted content, comments, liking behavior, frequency and nature of online interactions, serve as vital data sources for analyzing their psychological states. Research has uncovered correlations between specific behavioral patterns and linguistic habits on social media and depression, including frequent posting of negative content, increased nighttime activity, and reduced social engagement, all of which may be indicative of depressive symptoms\cite{vedula2017emotional}.For example, a study could analyze users' word choices in their texts, looking for frequent occurrences of negative emotion vocabulary like "loneliness," "fatigue," or "despair" as indicators of a depressive tendency\cite{fusar2023lived}. Moreover, the application of algorithms and machine learning technologies enables the processing and analysis of large-scale data, identifying more complex and subtle patterns that can facilitate earlier recognition of individuals likely experiencing depression by mental health professionals.In China, Sina Weibo, with its massive young user base, naturally emerges as a priceless resource for this kind of research. By scrutinizing the online behaviors of these users, researchers can not only track trends in mental health but also develop tools to facilitate early interventions.

Building upon prior work, in this paper, we aim to accomplish depression detection through user modeling with improved performance over previous efforts. To this end, we construct a new deep neural network classification model, the Multi-Modal Feature Fusion Model Using Cross-Attention. This model employs MacBERT\cite{cui2020revisiting} as the pre-training model to extract word features from text and incorporates an additional Transformer module to further refine context understanding specific to the task at hand, thereby enhancing adaptability to the targeted task. Departing from past practices, our approach utilizes Cross-Attention\cite{rajan2022cross} for multimodal feature integration instead of simply concatenating multiple features. Experimental proofs that our method effectively boosts the accuracy in detecting depression.

\textbf{Our Contributions.} The major contributions are listed as follows:
\begin{itemize}
 
\item We employ the Cross-Attention mechanism for fusing multimodal features, deviating from the common practice in prior works that typically resort to straightforward concatenation of multimodal attributes. The cross-attention strategy effectively captures and integrates these complementary pieces of information, enabling the model to conduct a more comprehensive and precise analysis of user emotions and behaviors. By calculating attention weights between features across different modalities, it highlights the most pertinent information from each modality. This capability of capturing associations is instrumental in enhancing the model's understanding and interpretation of relationships between distinct multimodal features.

\item We have constructed a deep neural network classification network, multimodal feature fusion network based on cross-attention(MFFNC), specifically designed for depression detection, capable of handling multimodal feature inputs. This model demonstrates exceptional performance in the task of depression identification. Our experiments have shown that compared to other prevalent classification models, ours achieves the highest level of accuracy and exhibits the greatest robustness.

\end{itemize}

\textbf{Paper Organization.} 
The remainder of this paper is organized as follows. Section~\ref{sec:background} introduces the background knowledge and elaborates on our research problem. Section~\ref{sec:approach} presents a detailed design of our model framework. Section~\ref{sec:implementation} demonstrates the implementation of our model. Section V reports the evaluation results of our model in predicting depression from social media content, along with comparisons to recent popular models. Section~\ref{sec:eval} summarizes the most relevant prior works. The final section, Section~\ref{sec:conclusion}, provides a conclusion for the entire paper.

\section{Background}
\label{sec:background} 
In this section, we introduce the dataset utilized in this paper as well as the relevant technologies employed.

\subsection{Weibo-User-Depression-Detection-Dataset(WU3D)}\label{sec:wu3d}
\begin{figure}[!htb]
  \includegraphics[width=0.48\textwidth]{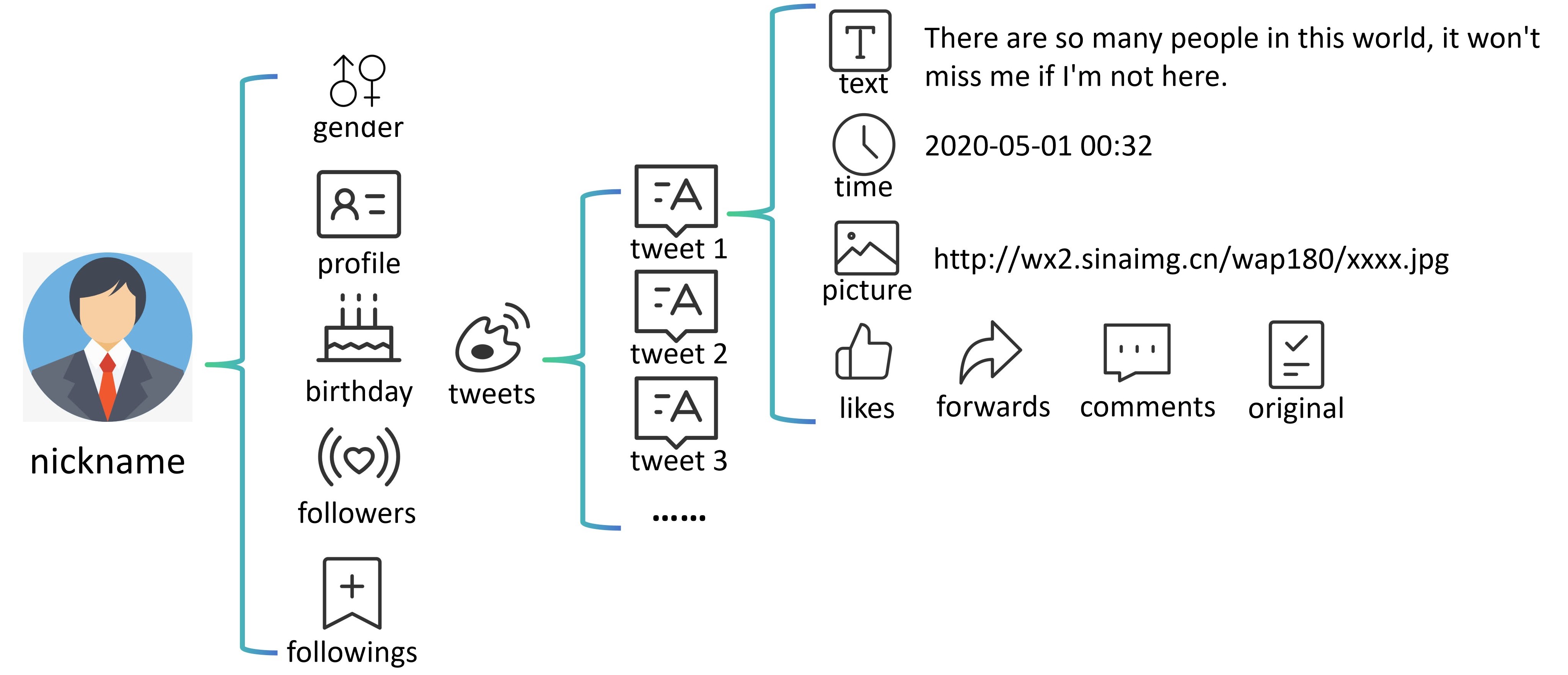}
  \caption{The data structure for each user in the WU3D comprises the following attributes: gender, profile information, birthday, the number of followers, the number of followings, and a collection of tweets. Each tweet within this collection includes details such as text content, posting time, presence of pictures, the number of likes received, the number of times it was forwarded, the number of comments, and an indicator specifying whether the tweet is an original post or a retweet.}
  \label{fig:wu3d}
\end{figure}

The dataset employed in our research is Weibo User Depression Detection Dataset (hereafter referred to as WU3D), compiled by Wang \emph{et al.}\cite{wang2020multimodal}. Specifically, within the Weibo platform, each user possesses a unique ID, and the WU3D dataset accesses the homepage of each user through web crawling techniques to gather information. With a focus on ensuring authority and high reliability, the dataset underwent dual scrutiny by psychologists and psychiatrists for labeling accuracy. The information collected for each user is illustrated in Fig.~\ref{fig:wu3d}.

\subsection{Attention}
Attention is a mechanism, which has found widespread application in Natural Language Processing (NLP) and computer vision tasks. At its core, the idea is to enable models to selectively focus on the most crucial parts of the input information during processing, rather than treating all input elements equally or averaging their importance. This mechanism mimics the way humans concentrate their attention on salient details when processing complex information, thereby enhancing the efficiency and performance of models.

\textbf{Transformer:} Transformer is a groundbreaking sequence transduction model, first proposed by Vaswani \emph{et al.} in their seminal paper 'Attention Is All You Need' in 2017\cite{vaswani2017attention}. It has utterly transformed the field of Natural Language Processing (NLP), reshaping the landscape dominated by Recurrent Neural Networks (RNNs) and their variants, such as Long Short-Term Memory (LSTM) and Gated Recurrent Units (GRUs). These models traditionally struggled with computational inefficiency when handling long sequences. Vaswani and colleagues addressed this by abandon the loop structure altogether, relying exclusively on self-attention mechanisms. This innovation demonstrated superior performance compared to the best RNN models of the time in machine translation tasks, while also significantly accelerating training speeds.

\textbf{Cross Attention:} Cross Attention is a specialized form of attention mechanism that primarily deals with dependencies between two different sequences\cite{crossattention}. Unlike self-attention, which focuses on interdependencies among elements within the same sequence, cross attention operates through a 'query-answer' paradigm, enabling one sequence to adjust its representation based on the content of another sequence. In this paper, to uncover the relationships between word features and statistical characteristics, the cross attention mechanism has been employed.

\subsection{Pre-trained Model}
Large-scale pre-trained models (PTMs), benefiting from intricate pre-training objectives and substantial parameter counts, excel at extracting knowledge from vast amounts of both labeled and unlabeled data. They encapsulate this abundant knowledge within their extensive parameters, which, when fine-tuned for particular tasks, can greatly enhance a wide array of downstream applications. This efficacy has been amply validated through rigorous experimentation and empirical assessments\cite{han2021pre}.

\textbf{MacBERT:} MacBERT\cite{cui2020revisiting} is an improved version of BERT, incorporating Error-Correcting Masked Language Model (MLM as Correction, abbreviated as Mac) into its pre-training regimen, addressing the inconsistency issue between "pre-training" and "downstream tasks." This model replaces the conventional MLM task with the MLM as Correction (Mac) task, thereby mitigating discrepancies between the pre-training and fine-tuning stages.

\section{Multimodal Feature Fusion Network based on Cross-attention}
\label{sec:approach} 
\begin{figure*}[!htb]
    \includegraphics[width=\textwidth]{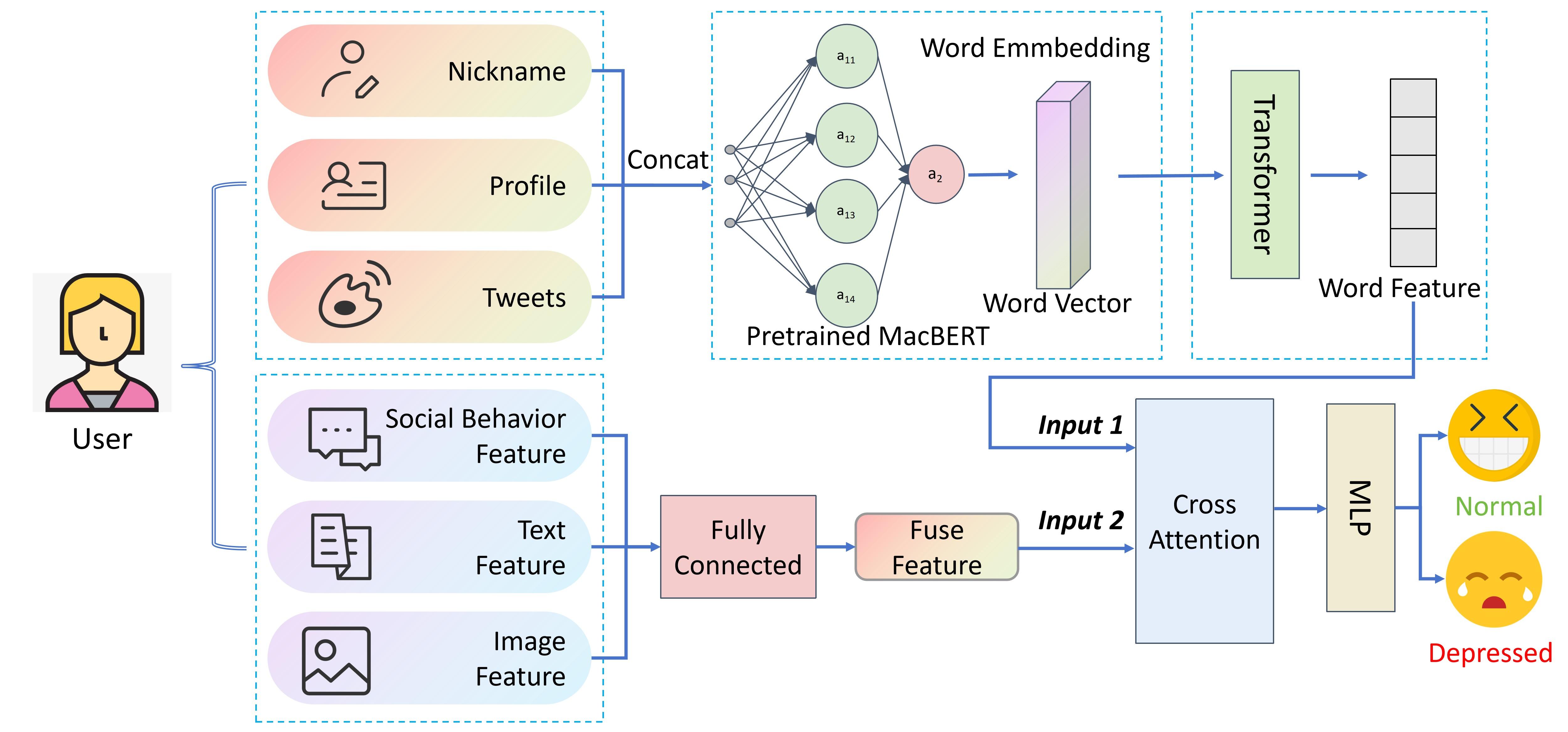}
    \caption{The Framework of the Network}
    \label{fig:network}
  \end{figure*}

This section describes the architecture of multimodal feature fusion network based on cross-attention(MFFNC), which comprises four fundamental components: word vector extracting, statistical feature extracting, feature fusion and multilayer perceptron(MLP), as shown in Fig.~\ref{fig:network}. The workflow commences by concatenating the user's nickname, profile, and tweets into a single, extended text sequence. This concatenated sequence is then fed into the pre-training model of MacBERT\cite{cui2020revisiting}, yielding embedded word vectors which serve as input 1. Concurrently, the text features, social behavioral features, and picture features are encoded separately to generate input 2. Subsequently, both input 1 and input 2 are channeled into a cross-attention module to perform cross-attention mechanisms, thereby yielding a fused feature representation. This fused feature is then propagated through a feedforward neural network for the detection of tendencies indicative of depression.

\subsection{Word Vector Extraction}
This module specializes in processing user text information originating from social media platforms, encompassing multiple dimensions of users, including nicknames, personal profiles, and posted tweets, among others. The process begins by integrating these various categories of information, concatenating them to form a continuous long-text sequence. This design is intended to capture and preserve the full picture of users' online behavior, as each component may harbor clues about their personality traits, emotional states, or social habits.

The constructed long-text sequence is then fed into a pre-trained MacBERT model. As an enhanced version of the BERT model, MacBERT introduces error-correcting masked language modeling tasks during pre-training, optimizing the model and reducing inconsistencies between pre-training and downstream tasks. This process enables not only the learning of lexical meanings but also the understanding of contextual nuances, extracting profound semantic information embedded within the text. Through MacBERT's processing, the raw text information is transformed into high-dimensional word embedding vectors. These vectors carry rich semantic and contextual information, furnishing high-quality input data, referred to as input 1, for subsequent natural language processing tasks.

This input 1, derived from deep learning model transformation, embodies a comprehensive understanding and parsing of user textual expressions. It not only facilitates the identification of key terms but also delves into the text to uncover emotional tendencies, underlying themes, and intricate patterns of interpersonal interactions. In applications such as depression detection, this nuanced text analysis capability is particularly crucial, enabling the recognition of subtle linguistic cues associated with depressive sentiments and providing robust data support for mental health assessments.

\subsection{Statistical Feature Extraction}
This module is responsible for extracting statistical information about users. Based on the research of previous work\cite{wang2020multimodal}\cite{shen2017depression}\cite{de2013predicting}\cite{shen2018cross}\cite{wang2013depression}\cite{lin2014user}\cite{cheng2017assessing}, we have adopted six statistical features: the proportion of negative emotional tweets, the proportion of original tweets, the ratio of posts made during late-night hours, the frequency of posts per week, the standard deviation of posting times, and the proportion of posts that include images. The detailed descriptions of these statistical features are provided in Table~\ref{tbl:statistic}. Detailed calculations for each of these statistical features along with their comprehensive descriptions will be elaborated upon in Section~\ref{subsec:statistic}. Following the manual extraction of these statistical features, they are fed into a fully connected network for encoding, serving as input 2, tailored to conform to the dimensional specifications required by the subsequent cross-attention mechanism module.

\begin{table*}[!htb]
    \caption{Statistical Feature}
    \label{tbl:statistic}
    \centering
    \begin{tabular}{c|l} 
    \hline
    \textbf{Feature} & \multicolumn{1}{c}{\textbf{Description}}                                                                                                                                                              \\ 
    \hline
    Social behavior  & \begin{tabular}[c]{@{}l@{}}The proportion of original tweets\\The proportion of late-night posts\\The frequency of posts per week\\The standard deviation of posting times\end{tabular}  \\ 
    \hline
    Text             & The proportion of negative emotional tweets                                                                                                                                                           \\ 
    \hline
    Image            & The frequency of image posting                                                                                                                                                           \\
    \hline
    \end{tabular}
\end{table*}

\subsection{Cross Attention}
Cross-attention\cite{gheini2021cross} is a mechanism predominantly used in Transformer models, calculating attention across two distinct sequences to manage semantic relationships between them. It finds applications in tasks such as machine translation, image captioning, and video-text alignment. This mechanism expands upon self-attention, enabling the model to dynamically aggregate information at each position of one sequence (the query sequence) based on the content of another sequence (the key-value sequence).

\begin{figure}[!htb]
    \includegraphics[width=0.48\textwidth]{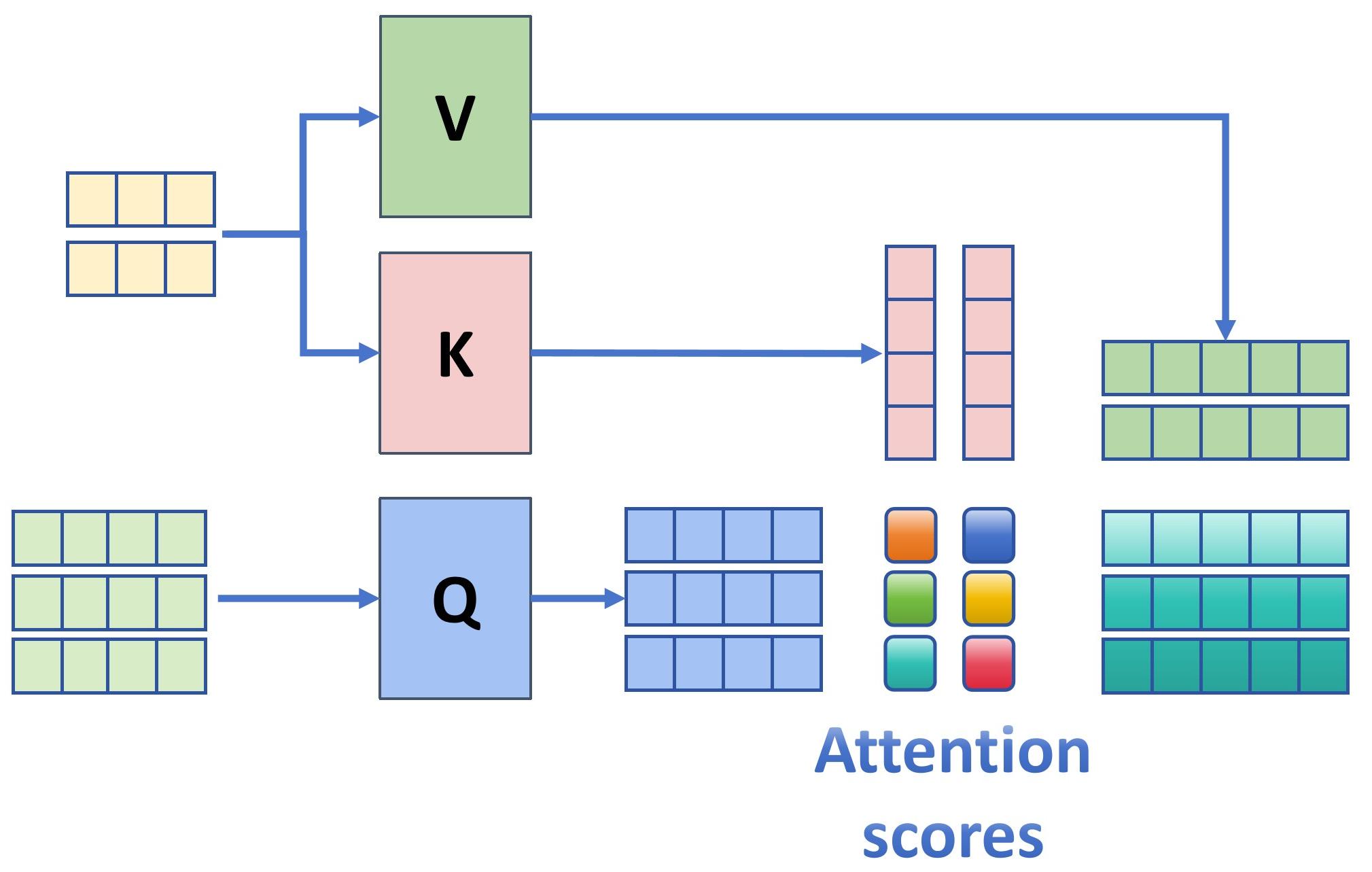}
    \caption{The Computation Process of Cross-attention}
    \label{fig:crossattention}
\end{figure}

The cross-attention mechanism is a specialized form of multi-head attention, where the input tensor is divided into two parts, $X_1\in\mathbb{R}^{n\times d_1}$ and $X_2\in\mathbb{R}^{n\times d_2}$.One part serves as the set of queries, while the other part acts as the key-value set.Its output is a tensor of size $n\times d_2$, where for each query vector, there are attention weights assigned to all the key vectors. Fig.~\ref{fig:crossattention} illustrates the computation process of cross-attention, where through matrix operations and the softmax function, the model learns to selectively gather pertinent information from the key-value sequences based on each element in the query sequence. This enhances the model's capability in handling cross-sequence dependencies, thereby boosting its performance. Such a mechanism is particularly crucial in dealing with multimodal tasks, as it enables the model to flexibly integrate features from disparate modalities. For instance, in the joint analysis of image and text, image features can serve as the key-value sequences, while the textual descriptions act as the query sequence. The cross-attention mechanism then facilitates precise alignment and information fusion between the two, thereby enhancing the interpretability and accuracy of the combined multimodal analysis.

Specifically, let $Q=X_{1}W^{Q}$ and $K=V=X_{2}W^{K}$,  the computation of the cross-attention is as follows:

\begin{equation}
  \label{equ:crossattention}
  \text{CrossAttention}(X_1,X_2)=\text{Softmax}\left(\frac{QK^T}{\sqrt{d_2}}\right)
\end{equation}
Where $W^Q\in\mathbb{R}^{d_1\times d_k}$ and $W^K\in\mathbb{R}^{d_2\times d_k}$ represent learned projection matrices,  $d_{k}$ denotes the dimensionality of the key-value set (which is also the dimensionality of the query set).

Below, we will elaborate on the computation process of cross-attention, as depicted in Figure 10086. This figure illustrates a mechanism known as "Cross-attention," a commonly employed technique in many modern Natural Language Processing (NLP) tasks. Within this process, V, K, and Q represent three matrices that collectively determine the attention scores. Here is a concise explanation of each component:

\begin{itemize}
  \item \textbf{V Matrix (Value):} The Value matrix encapsulates information from the input sequence, which has been encoded for the purpose of computing attention scores. Typically generated by an embedding layer of the input sequence, this layer is responsible for transforming raw text into numerical representations.
  \item \textbf{K Matrix (Key):} Similarly derived from the input sequence, the Key matrix serves to ascertain which positions are more significant during the calculation of attention scores. Elements in the K matrix are compared against those in the Q matrix to decide which positions ought to receive heightened attention.
  \item \textbf{Q Matrix (Query):} The Query matrix, another derivation from the input sequence, is utilized to inquire about the most relevant positions within the K matrix. Through similarity matching between elements in the Q matrix and those in the K matrix, it facilitates the determination of attention scores.
  \item \textbf{Attention Scores:} Attention scores are the outcome of the interaction between the V, K, and Q matrices. By assessing the similarity between the Q and K matrices, attention scores are computed. These scores are subsequently employed to weight and combine values from the V matrix, yielding the final output.
\end{itemize}

In practical applications, the cross-attention mechanism typically involves the following steps, as exemplified by Formula~\ref{equ:crossattention}:

\begin{enumerate}
  \item \textbf{Linear Transformation:} Initially, the V, K, and Q matrices undergo linear transformations, often through multiplication with weight matrices, to ensure they all possess the same dimensionality. This preprocessing aligns them for the subsequent computations.
  \item \textbf{Inner Product Operation:} Subsequently, each element in the Q matrix is paired with every element in the K matrix through an inner product operation, yielding a scalar value. These scalars quantify the degree of similarity between each element in the Q matrix and its corresponding elements in the K matrix.
  \item \textbf{Softmax Function:} The outcomes from the inner products are then fed into a softmax function. This operation scales the results such that each position's attention score falls within the interval $[0, 1]$, with the sum of all scores equaling 1. Consequently, a probability distribution is obtained, reflecting the relative importance of each position.
  \item \textbf{Weighted Summation:} Finally, the values in the V matrix are aggregated via a weighted summation process, where the weights are determined by the attention scores. This means that positions with higher attention scores contribute more significantly to the resultant output.
\end{enumerate}

The cross-attention mechanism is particularly valuable in neural network architectures like the Transformer, as it enables the model to focus on different segments of the input sequence. This is crucial for comprehending complex contextual dependencies and long-range relationships. Consequently, incorporating cross-attention as our multimodal feature fusion module proves highly efficient; it empowers the model to allocate attention resources flexibly, thereby achieving enhanced performance.

\subsection{Multi-Layer Perceptron}
The Multilayer Perceptron (MLP)\cite{taud2018multilayer}, short for Multilayer Perceptron, is a type of feedforward artificial neural network model that serves as an extended version of the single-layer perceptron. By incorporating one or more hidden layers, MLP enhances the learning capacity and expressive power of the network. This structural enhancement enables MLPs to fit complex nonlinear functional relationships, thereby excelling in solving many practical problems, particularly in classification and regression tasks.

In this paper, MLP is employed as the final classification head module, comprising two essential components: linear transformation layers and activation functions. Specifically, the MLP consists of two layers of linear transformations, with a Rectified Linear Unit (ReLU) activation function inserted between them. Renowned for its simplicity and efficacy, the ReLU activation function effectively alleviates the vanishing gradient problem, facilitating the training of deeper networks, as defined by the following equation:

\begin{equation}
\mathrm{MLP}(x)=\mathrm{ReLU}(W_2(\mathrm{ReLU}(W_1x+b_1))+b_2)
\end{equation}
where $\mathrm{MLP}(\cdot)$ represents the multilayer perceptron (MLP), $x$ represents the input feature vector, $W_{1}$ and $W_{2}$ are the weight matrices of the two linear transformation layers, and $b_{1}$ and $b_{2}$ respectively denote the bias terms for these two layers. The ReLU function is defined as $f(x)=\max(0,x)$, which maps all negative inputs to zero while preserving positive values unchanged. This approach maintains the nonlinearity of the network while circumventing the issue of gradient saturation in the negative value region.

Equipped with this architecture, the MLP is capable of learning high-level abstract features from the input data and, through its final output layer, provides category predictions. In the context of depression detection methodologies, the MLP serves as the decision layer, integrating multimodal features fused via the cross-attention mechanism. Through nonlinear transformations, these features are mapped onto a classification probability space, thereby facilitating accurate judgment regarding whether a user is experiencing depression. This design not only enhances the model's ability to recognize complex emotional and behavioral patterns but also ensures that the model exhibits good generalization performance, capable of making accurate predictions on unseen data.

\subsection{Psychological Analysis}
In this section, we delve into the inner world and behavioral manifestations of individuals with depression from a psychological perspective, to present readers with a vivid mental health landscape. Depression, a prevalent mental health issue worldwide\cite{mirowsky1992age}, not only impacts an individual's emotional experiences but also significantly alters their cognitive functions, social behaviors, and even physical health\cite{hammen2005stress}. Our aim, through meticulous data analysis, is to illuminate the distinctions between depression and other mental health states, thereby enhancing public comprehension of this complex psychological condition\cite{alexopoulos2005depression}.

\begin{enumerate}[label=\arabic*), start=1]
\item \textbf{Analysis of Characteristics in Individuals with Depression:}
The subjective experience of individuals with depression is often characterized by pervasive negative emotions, including sadness, hopelessness, self-deprecation, and loss of interest\cite{brown1977depression}. These emotions are not confined to specific situations but permeate throughout their daily lives, markedly diminishing their quality of life and sense of well-being. From a cognitive standpoint, those with depression frequently exhibit difficulties in concentrating, memory impairment, and diminished decision-making abilities – cognitive dysfunction that may be linked to alterations in the functioning of the prefrontal cortex and hippocampus regions of the brain. Furthermore, a pessimistic cognitive bias is prominent, manifesting as the overgeneralization of negative experiences and holding bleak outlooks for the future, which constitute key features of depression\cite{tiller2012depression}.
\item \textbf{Changes in Social Behavior:}
Social skills deficits are common among depressed patients, but little attention has been paid to this aspect of depression\cite{tse2004impact}. Within the realm of social interaction, individuals with depression tend to withdraw from their social circles and decrease engagement with others. Depressive symptoms can make individuals more sensitive to experiences of social exclusion and acceptance in their daily lives\cite{steger2009depression}. This retreat may stem from a lack of motivation to socialize, fear of being a burden to others, or an inability to derive pleasure from previously enjoyed social activities. Analysis of online behavior reveals that the social media usage patterns of those with depression also diverge; they often post more negative content, engage less in liking and commenting, and exhibit increased activity during late-night hours. These behavioral patterns, to some extent, mirror their psychological state.
\item \textbf{Comparison between the General Population and Individuals with Depression:}
In contrast, the general population exhibits a greater diversity in emotional expression, capable of experiencing and expressing a wide range of emotions, including positive ones such as joy and gratitude. Regarding cognitive function, they generally possess better control over attention, memory efficiency, and problem-solving capabilities. Socially, they tend to maintain stable social networks, engage in various social activities, and derive support and satisfaction from these interactions. Furthermore, their online behavior is more active and diverse, encompassing sharing daily life updates, participating in discussions, and fostering a broad network of social connections.
\end{enumerate}

\begin{figure}[!htb]
  \includegraphics[width=0.48\textwidth]{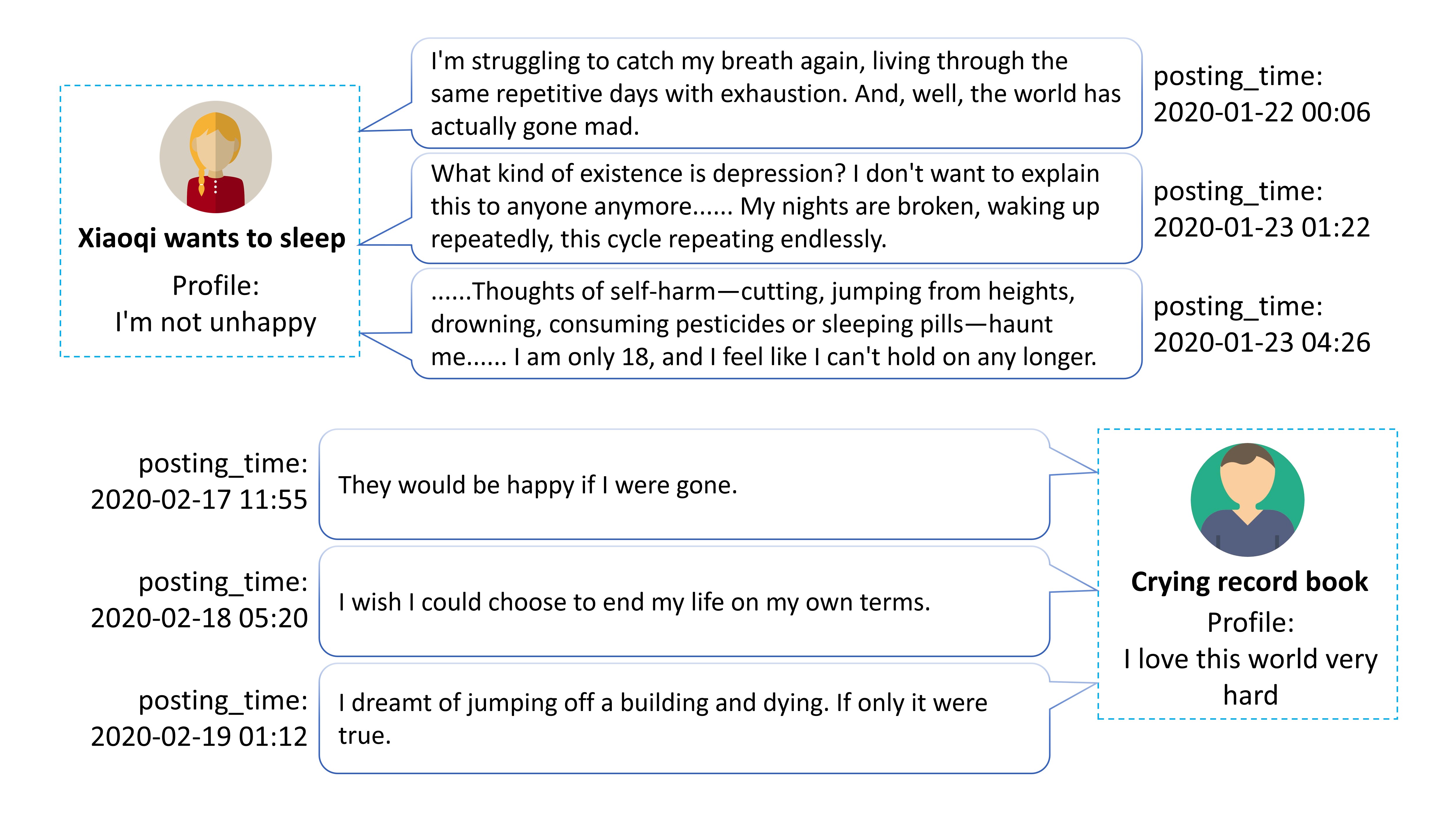}
  \caption{Case of patients with depression}
  \label{fig:case_d}
\end{figure}

According to Figure~\ref{fig:case_d}, we observe two social media users who have posted multiple tweets, which indicate they might be experiencing symptoms of depression. Let's analyze these tweets from a psychological perspective:

Firstly, the sentence, 'I'm struggling to catch my breath again, living through the same repetitive days with exhaustion. And, well, the world has actually gone mad.' suggests that the person (Xiaoqi wants to sleep) feels utterly exhausted, which can be a symptom of depression. Following this, 'What kind of existence is depression? I don't want to explain this to anyone anymore…… My nights are broken, waking up repeatedly, this cycle repeating endlessly,' indicates a loss of interest and pleasure in life, a common manifestation of depression. Furthermore, '……Thoughts of self-harm—cutting, jumping from heights, drowning, consuming pesticides or sleeping pills—haunt me……I am only 18, and I feel like I can't hold on any longer," reveals the presence of suicidal ideation, which is a severe indicator of a mental health crisis.

Another individual, referred to as ('Crying record book'), also expresses negative sentiments: 'They would be happy if I were gone.' And, 'I wish I could choose to end my life on my own terms.' These statements similarly demonstrate a sense of despair towards life and suicidal tendencies.

Taken together, these remarks all point towards the likelihood of depression. Depression is a treatable condition, and through our work, we are able to identify tendencies toward depression in its early stages, providing warning signals to professional caregivers to seek assistance, prompting intervention, and offering necessary psychological support or treatment recommendations. Timely intervention can improve the quality of life and prevent severe outcomes. Moreover, in extreme cases, it can avert tragedies such as suicide, saving precious lives.

\section{Implementation}
\label{sec:implementation}

In this section, we provide a detailed exposition on the implementation of the Multimodal Feature Fusion Network based on Cross-Attention (MFFNC), which was introduced in Section~\ref{sec:approach}. The overall architecture of the implementation is depicted in Fig.~\ref{fig:network}.

\subsection{Implementation of Word Vector Extraction}
In our study, to conduct a comprehensive analysis of users' emotional states, we integrate various pieces of information left by users on social media platforms, including their usernames, profile descriptions, and posts or retweets, forming a long text sequence. Notably, these posts comprise not only original content created by users but also retweets that shed light on their interests and emotional expressions. Through such integration, the constructed long text sequence transcends the limitations of individual posts, establishing strong contextual links across multiple tweets. Users may sequentially post tweets at different times to articulate their experiences battling depression or seek solace and express inner distress on social media. Consequently, aggregating this information is crucial for discerning whether a user is undergoing depression. We also observe that posts by depressed users do not always overtly convey distress or relate directly to depression; they may depict routine aspects of daily life. This necessitates meticulous semantic extraction from the text during depression prediction, distinguishing between diverse expressions and identifying subtle signs of depression amidst everyday content.

To address this, we employ the MacBERT model as an upstream tool for word embedding. MacBERT, an enhanced version of BERT, introduces a Masked Correction (Mac) pre-training task for the masked language model, resolving the inconsistency issue between pre-training and downstream tasks. By refining its profound understanding of linguistic contexts, MacBERT enhances performance in downstream tasks. Building upon this, we further leverage manually extracted statistical features, combining them with the word features output by MacBERT through a cross-attention module. The cross-attention mechanism effectively matches points of focus between two distinct feature sequences, fostering complementary information exchange and integration, thereby enhancing feature expressiveness.

Finally, the fused features are fed into a Multilayer Perceptron (MLP) classification model, acting as the decision-maker for the downstream task to yield the final depression detection outcome. Through this meticulously designed process, our system can more accurately identify indicators of depression from intricate social media data, providing robust technical support for early detection and intervention in mental health issues.

\subsection{Implementation of Statistical Feature Extraction}
\label{subsec:statistic}
Previous studies have defined features that are effective for detecting depressed users, we manually extracted six features, as detailed in Table~\ref{tbl:statistic}.

\subsubsection{The Proportion of Original Tweets}
The proportion of original tweets is a metric that measures the proportion of original content posted by users on a social media platform relative to retweeted content. This metric is particularly crucial in studies of user behavior, sentiment analysis, and especially in detecting mental health conditions such as depression, as it reflects users' tendencies for self-expression, their engagement in social interactions, and the authenticity of their emotions\cite{shen2018cross}. It calculates the percentage of original tweets posted by a user out of their total tweet count, which includes both original posts and retweets. Original tweets refer to those composed and shared directly by the user, rather than those forwarded from others. The computational formula is as follows:

\begin{equation}
  \text{Proportion of Original Tweets}=\frac{N_{original}}{N_{total}}\times100\%
\end{equation}
Where $N_{original}$ represents the number of original tweets posted by the user, $N_{total}$ denotes the total number of tweets by the user, including both original posts and retweets.

\subsubsection{The Proportion of Late-Night Posts}
The proportion of late-Night posts serves as a quantifiable metric reflecting users' social media activity patterns, playing a crucial role especially in research related to mental health. This indicator assesses the percentage of posts made during a defined "late-night" period out of an individual's total posts, offering insights into their sleep patterns, social interaction preferences, and potential mental well-being\cite{shen2017depression}. Although the exact timeframe for "late-night" may vary depending on the research context or cultural differences, it is commonly demarcated as the stretch from midnight to 6:00 AM. This period is chosen as it aligns with the human body's natural sleep cycle, and abnormal levels of activity during these hours can signal mood disorders, stress, insomnia, or other psychological distress.

The application of this metric extends beyond academic research and is increasingly being integrated into mental health monitoring systems, where it serves as one of the early indicators for identifying risks of mental health issues such as depression and anxiety. For instance, if a user consistently posts during late-night hours, this may signal the presence of sleep disorders or difficulties in emotion regulation\cite{wang2022people}, warranting further attention or intervention. The computational formula is as follows:

\begin{equation}
  \text{Proportion of Late-Night Posts}=\frac{N_{late-night}}{N_{total}}\times100\%
\end{equation}
Where $N_{late-night}$ is the number of posts made during late-night hours (midnight to 6 AM), $N_{total}$ is the total number of posts made during the entire observation period.

\subsubsection{The Frequency of Posts Per Week}
The frequency of posts per week is a metric that quantifies the average number of tweets a user posts within a week. This indicator sheds light on users' online behavioral patterns, levels of engagement, and potential trends in psychological well-being over time. In the realms of mental health and social behavior studies, both high and low frequencies of posting can be correlated with users' mental health status, highlighting its relevance for understanding psychological conditions through social media activity\cite{wang2020multimodal}. The computational formula is as follows:

\begin{equation}
  \text{Frequency of Posts Per Week}=\frac{T_{total}}{W_{period}}
\end{equation}
Where $T_{total}$ denotes the total number of posts made during the observation period, $W_{period}$ represents the number of weeks covered in the observation period (note that if the observation spans less than a week, adjust accordingly to the actual number of days covered and convert this into a fraction of a week, \emph{e.g.}, 3 days would be regarded as 0.43 weeks).

\subsubsection{The Standard Deviation of Posting Times}
The Standard Deviation of Posting Times is a metric that assesses the dispersion of users' posting time distribution. It reveals the degree of variation in posting timestamps throughout a day, indicating how scattered or concentrated their posting habits are. A larger standard deviation suggests a more dispersed pattern of posting times, potentially lacking a fixed or regular routine; conversely, a smaller standard deviation implies that users tend to post at more concentrated intervals, indicating a certain level of posting regularity. This measure provides insights into users' lifestyle habits, patterns of social behavior, and even mental health states, as irregular posting time distributions can sometimes correlate with sleep disorders or mood fluctuations. The computational formula is as follows:

\begin{equation}
  \text{SD} =\sqrt{\frac{\sum_{i=1}^n(x_i-\bar{x})^2}n}
\end{equation}
Where $X=\{x_1,x_2,...,x_n\}$ represents the set of posting times by the user during the observation period, $n$ denotes the total number of posts, $\bar{x}$ denotes the mean of the posting times.

\subsubsection{The Proportion of Negative Emotional Tweets}
In assessing the emotional tendencies and mental health status of social media users, accurately measuring the proportion of negative sentiment tweets is a crucial step. This endeavor capitalizes on the methodology proposed in \cite{wang2020multimodal}, ingeniously employing Baidu Intelligence Cloud's Sentiment Analysis API, a potent tool. Leveraging advanced natural language processing techniques and machine learning algorithms, this API delves into the emotional nuances of text, not only identifying basic positive or negative sentiments but also discerning varying degrees of negative emotion expressions, such as sadness, anger, or disappointment.

Implementation entails inputting each collected social media post as an individual data entry into the sentiment analysis system. The system then conducts a comprehensive scan of these contents, analyzing word choice, contextual frameworks, and underlying emotional cues. By thoroughly dissecting the text, the system outputs a proportion representing the intensity of negative sentiment within the content. As a result, both overt expressions of dissatisfaction and subtly implied pessimistic narratives are effectively identified and quantified.

This process is not only highly automated, enhancing the efficiency of data processing, but also ensures consistency and objectivity in assessment. The derived proportion of negative sentiment becomes a pivotal indicator of users' overall emotional state, particularly crucial in identifying those potentially in the early stages of depression or facing other mental health challenges. Through such quantitative analysis, researchers and mental health professionals can intervene earlier, providing necessary support and interventions to users, thereby promoting their mental health and wellbeing. Simultaneously, this analytical approach furnishes a reliable data foundation for large-scale mental health monitoring studies based on social media, propelling advancements in related scientific research.

\subsubsection{The Frequency of Image Posting}
When engaging in social activities, users often enrich their content by combining text with images, a multimodal communication strategy that significantly enhances the expressive power and appeal of the information conveyed. Images, as visual and vivid carriers of information, swiftly capture viewers' attention, conveying emotions and details difficult to express solely through text. Hence, posting pictures on social media has become a pivotal dimension in measuring users' content creation habits and patterns of social behavior.

The frequency of picture posts, or the proportion of posts containing images published by a user within a certain timeframe relative to their total posts, is a key indicator in assessing the level of visual content integration on social media platforms. This metric reflects an individual's preferred mode of communication with others: whether they are more inclined to utilize visually enriched formats to narrate stories, share snippets of their life, or express emotions and opinions. For researchers, analyzing picture posting frequency not only uncovers users' online behavioral tendencies, such as their preference for showcasing rather than merely narrating, but also indirectly provides insights into their personality traits, social involvement, and even mental health status\cite{kim2016predicting}. Users who frequently post images may prioritize visual experiences, seeking to connect with others in a more visual manner, which could relate to their outgoing nature and willingness to share. Conversely, those who post fewer images and rely more on text-based communication may have a stronger preference for deep thinking and textual expression, or they may be more conscious about privacy protection. Furthermore, studies on the social media behavior of groups prone to emotional lows, such as individuals with depression, indicate that they tend to be more active at night and share more negative content, yet variations in their image posting frequency can also reveal subtle differences in their emotional states; a reduction in image sharing might signify a decrease in social motivation. Therefore, through meticulous analysis of picture posting frequency, coupled with other social media behavioral statistical features, a multidimensional model of user behavior can be constructed. Such a model serves as a powerful tool for monitoring and preventing mental health issues in fields like public health and psychology research. The computational formula is as follows:

\begin{equation}
  \text{Frequency of Image Posting} =\frac{N_{image}}{N_{total}}
\end{equation}
where $N_{image}$ represents the number of tweets containing images, $N_{total}$ denotes the total number of tweets.
\section{Evaluation}
\label{sec:eval}

\subsection{Experiment Setup}
\subsubsection{Performance Metrics}
True Positive (TP), True Negative (TN), False Positive (FP), and False Negative (FN) form the foundation of the confusion matrix, which are metrics used to gauge the accuracy of model predictions. TP refers to instances where the model correctly predicted the positive class, TN where it correctly identified the negative class, FP represents cases where the model incorrectly labeled instances as positive, and FN denotes instances of the positive class that were incorrectly classified as negative. Based on these concepts, several common performance metrics can be defined to assess the model's performance:

\begin{equation}
  \text{Accuracy}=\frac{TP+TN}{TP+TN+FP+FN}
\end{equation}

\textbf{Accuracy}: Accuracy represents the proportion of all predictions that are correctly predicted.

\begin{equation}
  \mathrm{Recall}=\frac{TP}{TP+FN}
\end{equation}

\textbf{Recall}: Recall measures the model's ability to identify all true positive instances, that is, the proportion of actual positives that are correctly identified.

\begin{equation}
  \text{Precision}=\frac{TP}{TP+FP}
\end{equation}

\textbf{Precision}: Precision denotes the proportion of samples predicted as positive by the model that are actually positive, focusing on the accuracy of positive predictions.

\begin{equation}
  F1=2\times\frac{\text{Precision}\times\text{Recall}}{\text{Precision}+\text{Recall}}
\end{equation}

\textbf{F1-Score}: F1 Score ranges from 1 to 0, with a value closer to 1 indicating better model performance, particularly in scenarios where the distribution of positive and negative samples is imbalanced. The F1 Score serves as a more comprehensive evaluation metric under such conditions.

\subsubsection{Dataset}
For the datasets setup, we first divide our dataset by categorizing users into two groups: 'Depressed' and 'Normal', from which we sample 10,000 user data points from each category. Subsequently, we apply an 8:2 split ratio to this data (ensuring all experiments utilize a consistent random seed for sampling), resulting in a division of D1:D2 as 16,000: 4,000. This partitioning strategy ensures a balanced and reproducible allocation of data across training and Validation sets respectively.

\subsubsection{Environment Configuration}
The computational environment for this study is anchored within Alibaba Cloud's robust infrastructure—the Platform for Artificial Intelligence (PAI), specifically deploying a high performance interactive modeling tool known as the Data Science Workshop (DSW) server. This server boasts an exceptional configuration, equipped with 8 virtual CPU cores (vCPU), ensuring efficient multitasking and parallel processing capabilities. Its memory allocation stands at 30 GiB RAM, providing ample space for the instant loading of large datasets and the training of intricate models. Of particular note, the inclusion of an NVIDIA A10 GPU with 24GB of video memory lays a solid hardware foundation for the rapid training of deep learning models and the processing of massive data sets, especially suitable for computationally intensive tasks such as image recognition and natural language processing.

On the software front, the study embraces a cutting-edge technology stack to guarantee the project's sophistication and compatibility. Specifically, it utilizes the pytorch-develop version 2.1, a development iteration of the PyTorch deep learning framework that introduces the latest functional updates and performance enhancements, facilitating rapid model iteration and innovation. Python version 3.11, the most recent stable release at the time, was employed, having undergone substantial improvements in syntax sugars, performance, and type hinting, thereby augmenting code readability and execution efficiency. CUDA version 11.8, developed by NVIDIA as a parallel computing platform, ensures seamless integration between the GPU and deep learning frameworks, maximizing the utilization of GPU computational capabilities. The operating system selected was Ubuntu 22.04 LTS, renowned for its stability and a rich ecosystem of software, providing an advantageous environment for deep learning research.

To optimize model parameters, the Adam optimizer (Adaptive Moment Estimation) was adopted, an adaptive learning rate method lauded for its excellent convergence speed and generalization abilities across various tasks. A learning rate of $\text{1e-6}$ was set, a relatively conservative choice aimed at balancing model convergence speed with the mitigation of overfitting risks. Additionally, considering both efficiency and stability in model training, the batch size was determined to be 8, a common practice under limited hardware resources that strikes a balance between computational resource consumption and training effectiveness.

Regarding the loss function, CrossEntropyLoss was employed, a classic choice for multi-class classification problems. It effectively measures the discrepancy between the model's predicted probability distribution and the true labels, particularly when the last layer of a neural network outputs class probabilities, thereby enhancing the model's ability to learn more precise classification boundaries. Collectively, the computational environment and the implemented model strategies of this study embody a comprehensive consideration for efficiency, advancement, and practical feasibility, laying a sturdy groundwork for the smooth conduct of the research and the attainment of anticipated outcomes.

\subsection{Baseline Methods}
Not only do we compare against baseline methods, but we also incorporate several currently popular Chinese pre-trained models in combination with our feature fusion network, conducting separate experiments to evaluate the performance of our network.Furthermore, we investigate whether incorporating bi-GRU after the pre-trained model enhances the performance of our model.

\begin{enumerate}[label=\arabic*), start=1]
  \item XLNet with MFFN\cite{wang2020multimodal}:
  We reproduced their approach, as outlined in \cite{wang2020multimodal}, to establish our baseline. They adopted the widely-used pre-trained model, XLNet\cite{yang2019xlnet}, for extracting text-based word features. Subsequently, They designed a Deep Neural Network classification model, termed the Multimodal Feature Fusion Network (MFFN), which consolidates features from various information sources.

  \item LERT\cite{yiming2022lert}:
  LERT is a pre-training model that, in addition to employing masked language modeling, integrates three linguistic features and adopts a linguistically inspired pre-training mechanism. Experimental results demonstrate that the LERT algorithm can significantly enhance the performance of various pre-trained language models.

  \item PERT\cite{cui2022pert}:
  PERT is an innovative pre-training model that does not rely on the use of masked tokens ([MASK]) in conventional masked language models (MLMs) to learn semantic information from text. Instead, PERT adopts a methodology based on permuted language modeling to autonomously capture textual meaning. This approach involves rearranging the sequence of input text and then tasking the model with predicting the correct position of each token in the original text. Through this process, the model is compelled to learn deeper semantic structures and dependencies within sequences without explicitly concealing parts of the input.This method has demonstrated performance enhancements across a range of Chinese and English natural language understanding (NLU) tasks, indicating that pre-training through permuted language models effectively enhance the model's capability in understanding and generating text, paving the way for new pre-training techniques in the exploration of natural language processing.

\end{enumerate}

\subsection{Experimental Results}
\begin{figure*}[!htb]
  \includegraphics[width=\textwidth]{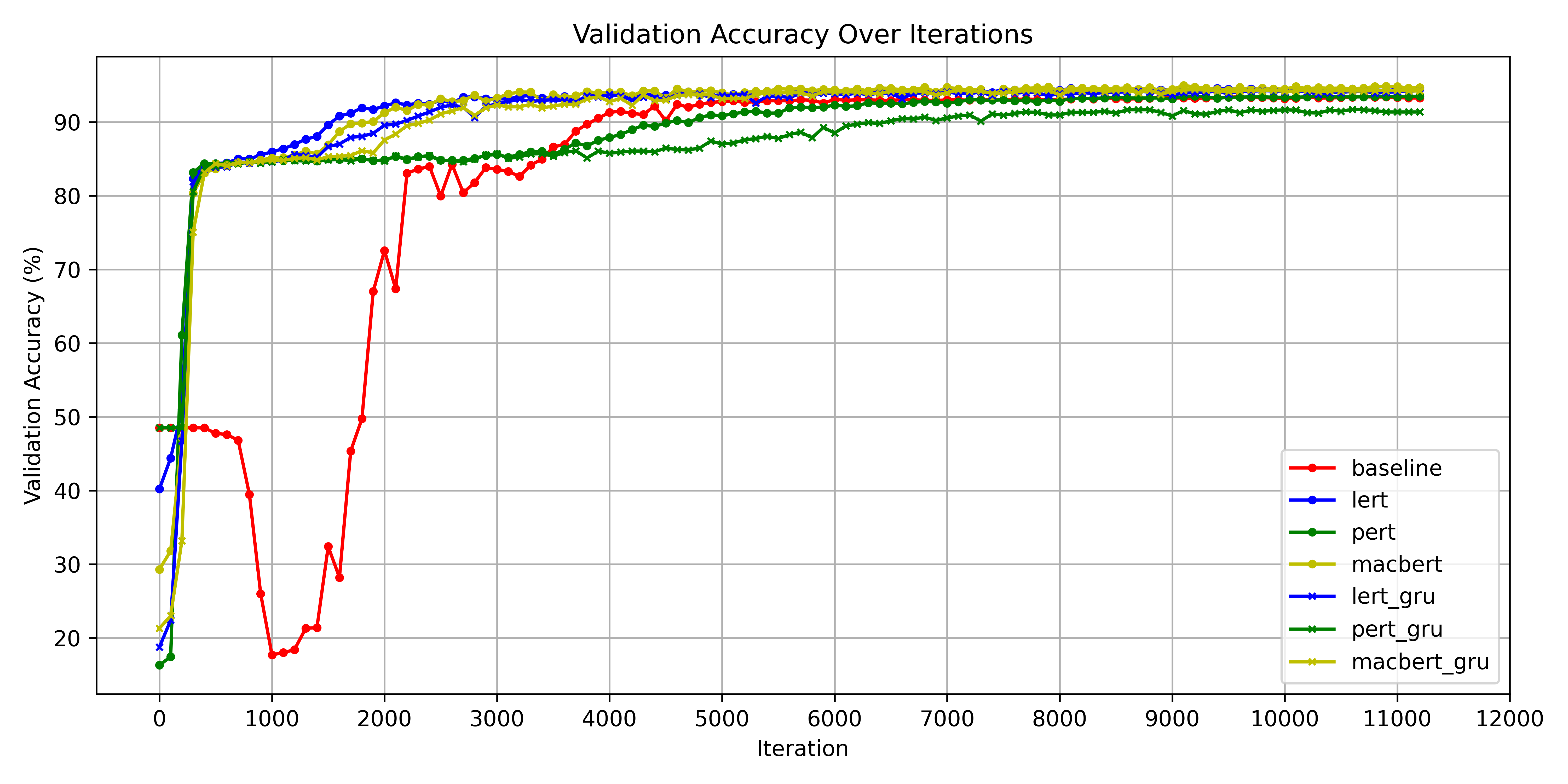}
  \caption{Validation Accuracy Over Iterations}
  \label{fig:result}
\end{figure*}

In this subsection, we provide a detailed analysis of the experimental results obtained by combining different pre-trained models with our cross-attention based feature fusion network.  Additionally, we examine the impact of incorporating bi-GRU following the pre-training model on the experimental results.The results are presented in Table~\ref{tbl:result} and Fig~\ref{fig:result}. Specifically, note that CA in the table denotes our method of fusing multimodal features using cross-attention.

\begin{table*}[!htb]
\caption{Experimental Results}
\label{tbl:result}
\centering
\begin{tabular}{ccccc}
\hline
\multirow{2}{*}{\textbf{Method}} & \multicolumn{2}{c}{\textbf{with bi-GRU}} & \multicolumn{2}{c}{\textbf{without bi-GRU}} \\ \cline{2-5} 
                                     & ACC                 & F1                 & ACC                  & F1                   \\ \hline
LERT+CA                          & 0.9425              & 0.9389             & 0.9455               & 0.9427               \\
PERT+CA                          & 0.9170              & 0.9101             & 0.9350               & 0.9298               \\
XLNET+MFFN                       & \multicolumn{4}{c}{ACC:0.9345 F1:0.9315}                                               \\
\textbf{MacBert+CA(our method)}  & \textbf{0.9445}     & \textbf{0.9413}    & \textbf{0.9495}      & \textbf{0.9469}      \\ \hline
\end{tabular}
\end{table*}

\begin{itemize}
\item Table~\ref{tbl:result} illustrates the experimental outcomes of various approaches, with and without the integration of Bidirectional Gated Recurrent Units (bi-GRU). Evident from the table, among all methods, MacBert+CA (our proposed method) excels in both accuracy and F1 score, irrespective of the inclusion of bi-GRU. This suggests that our method possesses strong generalization capabilities and robustness. Strikingly, however, for all methods, the addition of bi-GRU leads to a noticeable decrease in accuracy, leading us to conclude that bi-GRU has an adverse effect on these pre-trained models.

\item Fig.~\ref{fig:result} illustrates the evolution of validation set accuracy throughout the training process. Notably, apart from the baseline model, all other models incorporate our multimodal feature fusion approach, which employs cross-attention for integration and utilizes an MLP (Multi-Layer Perceptron) for the final classification task.  The baseline model (red curve) experiences a decline in accuracy during the early stages of training, but it unexpectedly begins to rise after Iteration 1000, eventually stabilizing at a level that, while consistent, remains lower than those of other enhanced models (with the exception of PERT+bi-GRU). Initially, the Lert (blue curve) and Pert (green curve) models exhibit similar performance; however, as training progresses, Lert surpasses Pert in terms of accuracy. Upon integrating the bi-GRU structure, as seen in Lert+bi-GRU (blue curve) and Pert+bi-GRU (another green curve), their accuracy rates generally decrease, yet both demonstrate robust stability throughout the training process.

\item Macbert (yellow curve) and Macbert+bi-GRU (yellow curve) emerge as the top performers among all models. Specifically, Macbert achieves high accuracy very early in the iteration phase and sustains this peak performance throughout the entire training period. The experiment suggests that adopting advanced techniques like Lert and Macbert can significantly enhance the validation accuracy of the model. Pert, on the other hand, performs marginally worse than other models in this task, even falling below the baseline accuracy at times, aligning with previous observations that its effectiveness can be task-dependent and caution is advised in its application.

\end{itemize}

\subsection{Ablation Study}
In order to comprehensively understand the impact of individual components within our proposed models, we conducted an ablation study. Specifically, we fed the word embeddings processed by MacBert into both bi-GRU and Transformer modules separately for feature extraction, thereby investigating their respective influences on the model's performance. The results of this ablation experiment are depicted in Table~\ref{tbl:ablation}.

The comparison between the bi-GRU and Transformer modules reveals that the Transformer module marginally outperforms the bi-GRU module. However, it is noteworthy that the standalone MacBert model generally excels over the MacBert models augmented with either the bi-GRU or Transformer modules, affirming the efficacy of MacBert in this particular task. The performance of the MacBert network with the Transformer module added and without any additional module is nearly equivalent, exhibiting accuracies (ACC) of 0.9495 and 0.9490, and F1-scores of 0.9469 and 0.9465, respectively. Our ablation study underscores the significant role each component plays in the holistic performance of the model.

\begin{table}[!htb]
\caption{Ablation Study}
\label{tbl:ablation}
\centering
\begin{tabular}{ccc}
\hline
\textbf{Method}        & \textbf{ACC} & \textbf{F1} \\ \hline
MacBert+CA             & 0.9495       & 0.9469      \\
MacBert+bi-GRU+CA      & 0.9445       & 0.9413      \\
MacBert+Transformer+CA & 0.9490       & 0.9465      \\ \hline
\end{tabular}
\end{table}

\begin{table*}[!htb]
  \caption{Case 1}
  \label{tbl:case1}
  \centering
  \begin{tabularx}{\textwidth}{l X c}
    \toprule
    \textbf{User} & \textbf{Tweets} & \textbf{Depression/Normal} \\
    \midrule
    \multirow{2}{*}{\begin{tabular}[c]{@{}l@{}}Nickname:\\ Director Ren Youbing \end{tabular}} 
    & Am I a bit depressed? Should I take some medication? \\ \cmidrule(lr){2-2}  
    & Depression is creeping up on me slowly. Every day, I find it hard to feel happy; it's truly uncomfortable. Indeed, the environment does shape one's mood. When I leave, my spirits seem to depart with me...
    & \multirow{4}{*}{Depression} \\ \cmidrule(lr){2-2}
    \addlinespace
    \multirow{2}{*}{\begin{tabular}[c]{@{}l@{}}Profile:\\ Director of the depression \\ research institute \end{tabular}}
    & I've posed the question: What are some self-healing methods for mild depression? \\ \cmidrule(lr){2-2}  
    & \ldots 
    & \\
    \bottomrule
  \end{tabularx}
\end{table*}

\begin{table*}[!htb]
  \caption{Case 2}
  \label{tbl:case2}
  \centering
  \begin{tabularx}{\textwidth}{l X c}
    \toprule
    \textbf{User} & \textbf{Tweets} & \textbf{Depression/Normal} \\
    \midrule
    \multirow{2}{*}{\begin{tabular}[c]{@{}l@{}}Nickname:\\ Mr. Yusan is also begging \\ for him today\end{tabular}} 
    & After two days and nights of intense pain, my little angel was born - proof that beauty arises from hardship. \\ \cmidrule(lr){2-2}  
    & Deadline pressing, all-nighters can't compete with senior scholars. Yet, the scenery outside reminds me of life's beauty beyond this rush.
    & \multirow{4}{*}{Depression} \\ \cmidrule(lr){2-2}
    \addlinespace
    \multirow{2}{*}{\begin{tabular}[c]{@{}l@{}}Profile:\\ We can make it right $\sim$\end{tabular}}
    & I think I might be depressed. I've been feeling conflicted all day and have a strong urge to die. What should I do? \\ \cmidrule(lr){2-2}  
    & \ldots 
    & \\
    \bottomrule
  \end{tabularx}
\end{table*}

\begin{table*}[!htb]
  \caption{Case 3}
  \label{tbl:case3}
  \centering
  \begin{tabularx}{\textwidth}{l X c}
    \toprule
    \textbf{User} & \textbf{Tweets} & \textbf{Depression/Normal} \\
    \midrule
    \multirow{2}{*}{\begin{tabular}[c]{@{}l@{}}Nickname:\\ Cloncurry \end{tabular}} 
    & I really can't stand the sound of teeth grinding. It keeps me awake and now I'm hungry too. \\ \cmidrule(lr){2-2}  
    & Meet the best version of yourself in the most beautiful time. Let's go, we've got this!
    & \multirow{4}{*}{Normal} \\ \cmidrule(lr){2-2}
    \addlinespace
    \multirow{2}{*}{\begin{tabular}[c]{@{}l@{}}Profile:\\ Calmness comes first \\ from within oneself. \end{tabular}}
    & I entrust my youth here, where it will shimmer brightly. Starting today, I strive to forge ahead with all my might. \\ \cmidrule(lr){2-2}  
    & \ldots 
    & \\
    \bottomrule
  \end{tabularx}
\end{table*}

\subsection{Case Study}
We conducted a case study to explore the accuracy and robustness of our model.

We have provided three examples for analysis, as shown in Table~\ref{tbl:case1}, Table~\ref{tbl:case2}, and Table~\ref{tbl:case3}. Let us proceed to analyze their respective outcomes:

\begin{enumerate}
  \item Case 1:  
  Table~\ref{tbl:case1} presents the profile information and tweets of a user named 'Director Ren Youbing' who identifies himself as the 'Director of the Depression Research Institute.' Through his profile description and the content of the illustrated tweets, signs of mild depression are noticeable. Remarkably, 'Director Ren Youbing' exhibits a degree of self-awareness, openly acknowledging his mental health state. In one particular tweet, he inquires, 'I've posed the question: What are some self-healing methods for mild depression?' This demonstrates not only his awareness of his condition but also a proactive approach in seeking remedies through social media platforms.
  
  For users like 'Director Ren Youbing', whose profiles and tweets frequently incorporate negative terms such as 'depression,' our model demonstrates a high level of sensitivity and precision. It is capable of accurately identifying depressive tendencies and swiftly converging on predictions, highlighting the efficacy and reliability of our model in assessing mental health indicators within online contexts.
  
  \item Case 2: 
  Table~\ref{tbl:case2} introduces an individual with the username 'Mr. Yusan is also begging for him today,' showcasing how relying solely on a limited number of tweets for psychological profiling can lead to misinterpretations. The user's nickname, profile description, and selected tweets illustrate this point. Particularly, the user's profile statement, 'We can make it right $\sim$,' embodies a profound sense of optimism, powerfully expressing the user's positive attitude and belief in the face of life's challenges. The profile summary, in its concise yet potent form, not only outlines the user's basic profile but, in this case, subtly reveals their inherent resilience and determination to improve their circumstances. This further adds to the complexity of understanding their psychological state.
  
  For instance, the tweet, 'After two days and nights of excruciating pain, my little angel was born – a testament to beauty blossoming amidst agony,' reveals that the user is actually a new mother rejoicing over the arrival of her child after the ordeal of childbirth. Another tweet, 'With the deadline looming, burning the midnight oil seems inadequate compared to seasoned scholars' composure, yet the view outside my window reminds me that amidst this rush, the beauty of life still deserves appreciation," demonstrates that despite intense work pressure, the user finds solace in natural scenery, transforming stress into motivation. Terms like 'little angel' and references to 'beauty' convey an upbeat sentiment.

  Contrastingly, another tweet states, 'I fear I might be suffering from depression, plagued by inner turmoil all day, even entertaining strong suicidal thoughts. What should I do?' This clearly expresses negative emotions and psychological distress. By comprehensively analyzing the content and varied emotional hues of this user's tweets, our model not only detects direct cues like 'depression' but also uncovers underlying depressive tendencies from seemingly positive expressions. Through integrating multi-faceted information, our model achieves a more precise assessment. This capability highlights the robustness of our model, which maintains a high accuracy in identifying depressive tendencies amidst the complexity and variability of social media text data.
  
  \item Case 3: 
  In contrast to the cases presented in Table~\ref{tbl:case1} and Table~\ref{tbl:case2}, Table~\ref{tbl:case3} showcases a user profile and tweet samples of an individual whom our model predicts to have a “normal” mental state. While one of the user’s tweets, "I really can’t stand the sound of teeth grinding. It keeps me awake and now I’m hungry too," hints at experiencing insomnia, particularly when coupled with the statistical feature that most of their tweets are posted late at night, this initial observation may suggest a preliminary suspicion of depressive tendencies.

  However, upon deeper analysis of subsequent tweets, we find that the user, in fact, harbors an optimistic outlook towards the future, filled with anticipation and vision for life. This stark contrast underscores the superficiality of initial judgments. This process further validates the robustness of our model—it refrains from fixating on singular pieces of information or momentary emotional expressions. Instead, it holistically evaluates the user's diverse tweets across different times and contexts, conducting a meticulous, multidimensional assessment that significantly reduces the risk of misclassification. This example affirms the model's precision and profound understanding in the realm of complex emotion recognition and mental health assessment.
\end{enumerate}
\section{Related Works}
\label{sec:related}
With the advancement of artificial intelligence technology, machine learning approaches have made significant contributions to depression detection on social media. These efforts can be categorized into two main classes: traditional machine learning methods and deep learning-based natural language processing techniques.

\subsection{Traditional Machine Learning Approaches}
Early studies, such as those by Choudhury and colleagues\cite{de2013predicting}, employed an elaborate manual feature engineering process to extract features from Twitter users' behavioral patterns, providing inputs for subsequent depression detection models. This approach relies on researchers' prior knowledge of depressive manifestations to manually design features that reflect depressive tendencies. Li \emph{et al.}\cite{li2020impact} utilized sentiment analysis techniques, in conjunction with word embedding methods, to assess the depressive tendency of tweets. This underscores the importance of textual content, particularly emotional expression, in identifying depressive states, further affirming the pivotal role of text-based features in online mental health analyses. Shen \emph{et al.}\cite{shen2017depression}, in their research, considered not only textual information but also integrated other data types (such as user behavior data), adopting multimodal methods to construct a more comprehensive user profile. This indicates that incorporating diverse information sources can effectively enhance the accuracy and robustness of depression detection. 

\subsection{Natural Language Processing Detection Approaches based on Deep Learning}
Recent research has shifted towards employing deep learning, notably neural network models, for processing social media data and identifying depressive users. The pioneering work by Mustafa and colleagues\cite{mustafa2020multiclass} marks a transition in the field of online social network (OSN) depression detection, shifting from conventional machine learning to the deep learning paradigm. This transition showcases the capability of automatically learning complex and high-dimensional features, thereby enhancing both the efficiency and accuracy of detection.Lin \emph{et al.}'s work\cite{lin2020sensemood} illustrates how cross-disciplinary technological innovation, specifically the integration of state-of-the-art NLP pre-traind models with image processing techniques, can advance mental health monitoring. They employed BERT (Bidirectional Encoder Representations from Transformers)\cite{devlin2018bert}, a groundbreaking pre-trained model, which through deep bidirectional context understanding, generates high-quality word embeddings (i.e., high-dimensional vectors) that capture the nuanced contextual meanings of words in sentences, vastly outperforming previous bag-of-words models or simplistic word embedding approaches.Specifically, Wang \emph{et al.}\cite{wang2020multimodal} constructed a dataset, the Weibo User Depression Detection Dataset (WU3D), and employed the popular pre-trained model XLNet to extract text-based word features. They then devised a Deep Neural Network classification model, the Multimodal Feature Fusion Network (MFFN), which integrates features derived from diverse information sources, further accomplishing the classification task. Their approach demonstrated remarkable performance on the test dataset. The dataset utilized in this paper is the publicly available WU3D dataset provided by their team.

\section{Conclusions}
\label{sec:conclusion}
This paper presents a depression detection method based on multi-modal feature fusion using cross-attention. By employing MacBERT as the pre-training model and incorporating the cross-attention mechanism, we have significantly enhanced the accuracy of depression detection. Experimental outcomes demonstrate that our approach surpasses other models, achieving an accuracy of 0.9495 and an F1 score of 0.9469. In comparison to alternative methods, ours effectively integrates multi-modal features, thereby furnishing a more precise depression detection. In summary, our method represents a significant advancement in depression detection from social media, offering a potent and accurate tool for mental health monitoring. Future work may explore the integration of additional modalities, such as visual and audio data, to further enhance detection capabilities.

Of paramount importance is the core value of our research, which lies in the potential to integrate this model seamlessly into mainstream social media platforms like Weibo or WeChat. By doing so, we can facilitate accurate early diagnosis and warning signs of depression at its nascent stage. This real-time intervention mechanism provides a crucial window for initiating psychological interventions promptly, effectively bridging a pathway of hope for those struggling with depression. Empowering mental health care through technology, we aspire to weave a safety net in the digital era, extending care and support to every corner where hearts in need reside.

\bibliographystyle{IEEEtran}
\bibliography{paper}

\end{document}